\documentclass[journal]{IEEEtran}

\usepackage{amsmath,amssymb,amsfonts}
\usepackage{graphicx}
\usepackage{url}
\usepackage{cite}
\usepackage{tikz}
\usepackage{algorithmic}
\usepackage{algorithm}
\usepackage{multirow}
\usepackage{pifont}
\usepackage{microtype}
\usepackage{siunitx}
\usepackage{booktabs} 
\usepackage[justification=centering]{caption}
\usepackage[font=small,labelfont=bf]{caption}
\usepackage{caption}
\usepackage[table]{xcolor} 
\usepackage{xcolor}
\usepackage{colortbl}

\usepackage{booktabs}
\usepackage{multirow}
\usepackage{tabularx}
\usepackage{array}

\usepackage{booktabs}
\usepackage{xcolor}
\usepackage{colortbl}
\usepackage{makecell}
\usepackage{graphicx}
\usepackage{orcidlink}
\usepackage{caption}
\title{
OneViewAll: Model-Free Single-View 6D Pose Estimation via Projection-based Alignment}

\author{
    Yang Luo\orcidlink{0009-0002-7124-2296}, 
    Yan Gong\orcidlink{0000-0002-3148-8286}, 
    Yongsheng Gao*\orcidlink{0000-0002-1555-8328}, 
    Jie Zhao\orcidlink{0000-0002-6086-9387},~\IEEEmembership{Senior Member,~IEEE}, 
    Xinyu Zhang\orcidlink{0000-0003-0034-9037},~\IEEEmembership{Member,~IEEE}, 
    and Huaping Liu\orcidlink{0000-0002-4042-6044},~\IEEEmembership{Fellow,~IEEE}%

\thanks{This work was supported by the National Science and Technology Major Project (Grant No. 2025ZD1603200) and the National Outstanding Youth Science Fund of the National Natural Science Foundation of China (Grant No. 52025054). (\textit{Corresponding author: Yongsheng Gao})}%
\thanks{Yang Luo, Yan Gong, Yongsheng Gao, and Jie Zhao are with the State Key Laboratory of Robotics and Systems, Harbin Institute of Technology, Harbin 150001, China (email: christoluo@outlook.com; gongyan2020@foxmail.com; gaoys@hit.edu.cn; jzhao@hit.edu.cn).}%
\thanks{Xinyu~Zhang is with the State Key Laboratory of Intelligent Green Vehicle and Mobility, the School of Vehicle and Mobility, Tsinghua University, Beijing 100084, China (email: xyzhang@tsinghua.edu.cn).}%
\thanks{Huaping~Liu is with the Department of Computer Science and Technology, Tsinghua University, Beijing 100084, China (email: hpliu@tsinghua.edu.cn).}%
}

\begin{document}

\maketitle

\begin{abstract}
In many practical 6D object pose estimation scenarios, only a single real-world RGB-D reference view per object is available, without access to CAD models. Most existing high-accuracy methods, however, rely on explicit 3D models or multi-view data, which limits their scalability and deployment. We propose OneViewAll, a projection-based framework that estimates 6D pose from a single RGB-D reference without CAD models or explicit 3D reconstruction. Instead of rendering CAD models, OneViewAll generates pose-conditioned projections directly from the reference observation and aligns them with the query through iterative refinement. To handle the incompleteness of single-view observations, we introduce a self-checked symmetry hypothesis that hallucinates plausible back-side geometry via mirror fusion, providing a geometric proxy for improved robustness under large viewpoint changes and self-occlusions. In addition, our methods optionally prune implausible hypotheses during initialization, improving the accuracy-efficiency trade-off. Experiments demonstrate that OneViewAll achieves \textbf{91.8\%} ADD-0.1d accuracy on LINEMOD with a single real-world reference view, and delivers performance gains on LM-O, Real275, and Toyota-Light. Code is available at: \url{https://github.com/tilaba/OneViewAll.git}.
\end{abstract}


\section{Introduction}
6D object pose estimation aims to recover an object's rotation and translation relative to the camera~\cite{krishnan2024ominnocs, UA-Pose_2025}, thereby serving as a cornerstone for applications such as robotic grasping~\cite{CaTGrasp_2022} and virtual reality~\cite{Hands_on_2016}. Despite significant progress driven by deep learning and 3D vision, existing methods often rely on high-quality CAD models~\cite{FoundationPose_2024, SAM6D_2024, GigaPose_2024} or multi-view data with accurate annotations~\cite{liu2022gen6d, OnePose_2022}, which are costly and impractical in real-world scenarios. This raises a critical challenge: how to achieve accurate and efficient 6D pose estimation with minimal reference information, ideally from only a single reference view together with a query image.

Most high-accuracy methods on benchmarks such as BOP~\cite{FoundationPose_2024, co-op_2025, Freeze_v2} fall into two categories: correspondence-based and render-and-compare. The former establishes 2D--3D correspondences between image observations and CAD models~\cite{Self6D_2024, Foundpose_2024, Corr2Distrib_2025}, followed by pose estimation via PnP~\cite{lepetit2009epnp} and RANSAC~\cite{fischler1981ransac}. The latter synthesizes candidate views from a CAD model and compares them with the query image via image or feature matching~\cite{DeepIM_2018, labbe2020cosypose}. While render-and-compare methods often achieve stronger robustness under occlusion and challenging lighting~\cite{ bop2024challenge}, the prerequisite of a high-fidelity CAD model forms the primary barrier: acquiring such models is expensive and labor-intensive, severely limiting scalability and real-world deployment.

\begin{figure}
    \centering
    \includegraphics[trim={1cm 1.5cm 7.2cm 2cm}, clip, width=1\linewidth]{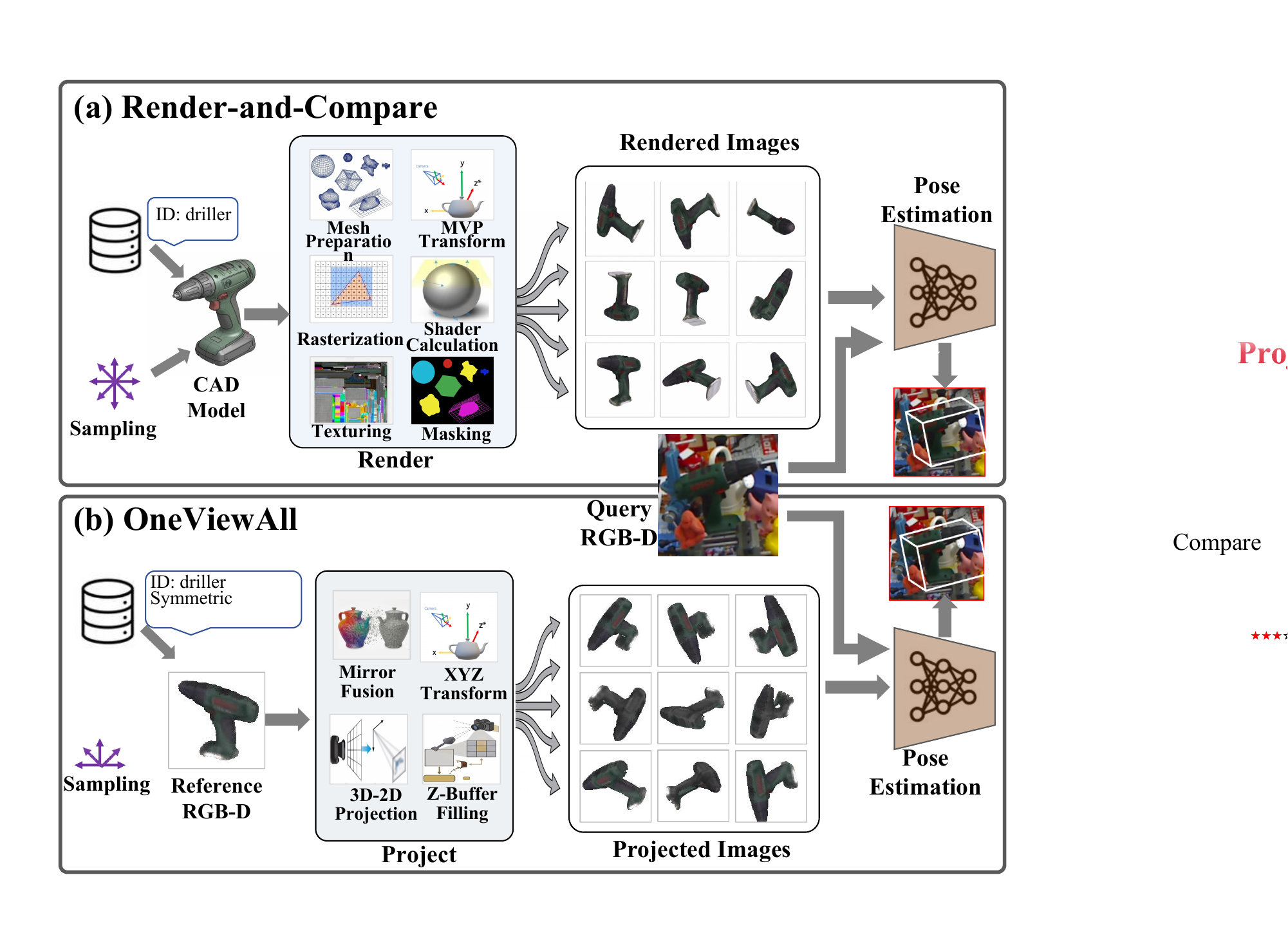}
    \captionsetup{
        font=small,
        labelfont=bf,
        justification=justified,
    }

    \caption{Render-and-Compare vs. OneViewAll for model-free 6D pose estimation. (a) Traditional render-and-compare relies on CAD models and rendering for hypothesis generation and comparison. (b) OneViewAll operates on a single reference RGB-D view using symmetry-aware projection, enabling efficient pose alignment without explicit 3D assets or multi-view data.}
    \label{fig:comparison}
\end{figure}

In response, recent studies~\cite{NOPE_2024, One2Any_2025, Any6D_2025, Liu2025SinRef6D} have explored approaches that require no CAD model at inference time (often referred to as model‑free). Despite removing explicit CAD dependency, these methods still rely on either geometric reconstruction or data-intensive learning paradigms, and often suffer from high computational cost or limited generalization across unseen objects and scenes.

Compared with multi-view setups, single-reference-view model-free 6D pose estimation~\cite{One2Any_2025, Any6D_2025, Liu2025SinRef6D} is more practical yet more challenging. While One2Any~\cite{One2Any_2025} established this setting, its performance remains limited under challenging viewpoints. SinRef-6D~\cite{Liu2025SinRef6D} improves results via geometric alignment, though it struggles with large viewpoint variations. Alternatively, Any6D~\cite{Any6D_2025} enhances geometric completeness via explicit 3D representations for render-and-compare, but it is computationally expensive and its pose accuracy heavily depends on the reconstructed CAD quality, which is complex to obtain.

Although single-reference model-free methods have shown promising performance, they still suffer from three major limitations: (1) reliance on explicit 3D reconstruction or multi-view rendering, which leads to high computational overhead and limited inference accuracy; (2) limited robustness to large viewpoint changes and severe self-occlusions; and (3) unresolved global pose ambiguity, especially for symmetric or texture-less objects.

To address these challenges under the strict single-reference setting, we propose OneViewAll, a Project-and-Compare framework that estimates 6D pose from a single RGB-D reference view without CAD models or 3D reconstruction. OneViewAll warps and splats the reference observation onto candidate viewpoints through pose-conditioned projection, and refines the pose by comparing these projections with the query. Unlike rendering which demands a complete 3D model, this projection operates on the raw reference image directly and preserves the original geometric fidelity, eliminating the need for intermediate geometry and texture reconstruction, as shown in Fig.~\ref{fig:comparison}.
Additionally, OneViewAll hypothesizes potential object symmetries and verifies their consistency with the reference view. The plausible symmetry is then applied in mirror fusion to infer invisible back-side geometry, providing extra cues for pose alignment under large viewpoint changes and self-occlusions. The main contributions of this work are summarized as follows:
\begin{itemize}
\item \textbf{Project-and-Compare framework:} We introduce a projection-based pose estimation pipeline that operates from a single RGB-D reference view, without requiring CAD models, explicit 3D reconstruction, or multi-view rendering.

\item \textbf{Symmetry-guided mirror fusion:} We introduce a self-check procedure that selects a plausible symmetry plane from the reference view and hallucinates invisible back-side geometry via mirror fusion, providing a geometric proxy for robust pose alignment under large viewpoint changes and occlusions.

\item \textbf{Accuracy-efficiency advantage:} Experiments demonstrate that OneViewAll achieves state-of-the-art performance, including 91.8\% ADD-0.1d accuracy on LINEMOD using real reference views.
\end{itemize}

\begin{figure*}[t]
    \centering
    \includegraphics[trim={2.6cm 4cm 4.75cm 5.2cm}, clip, width=1\linewidth]{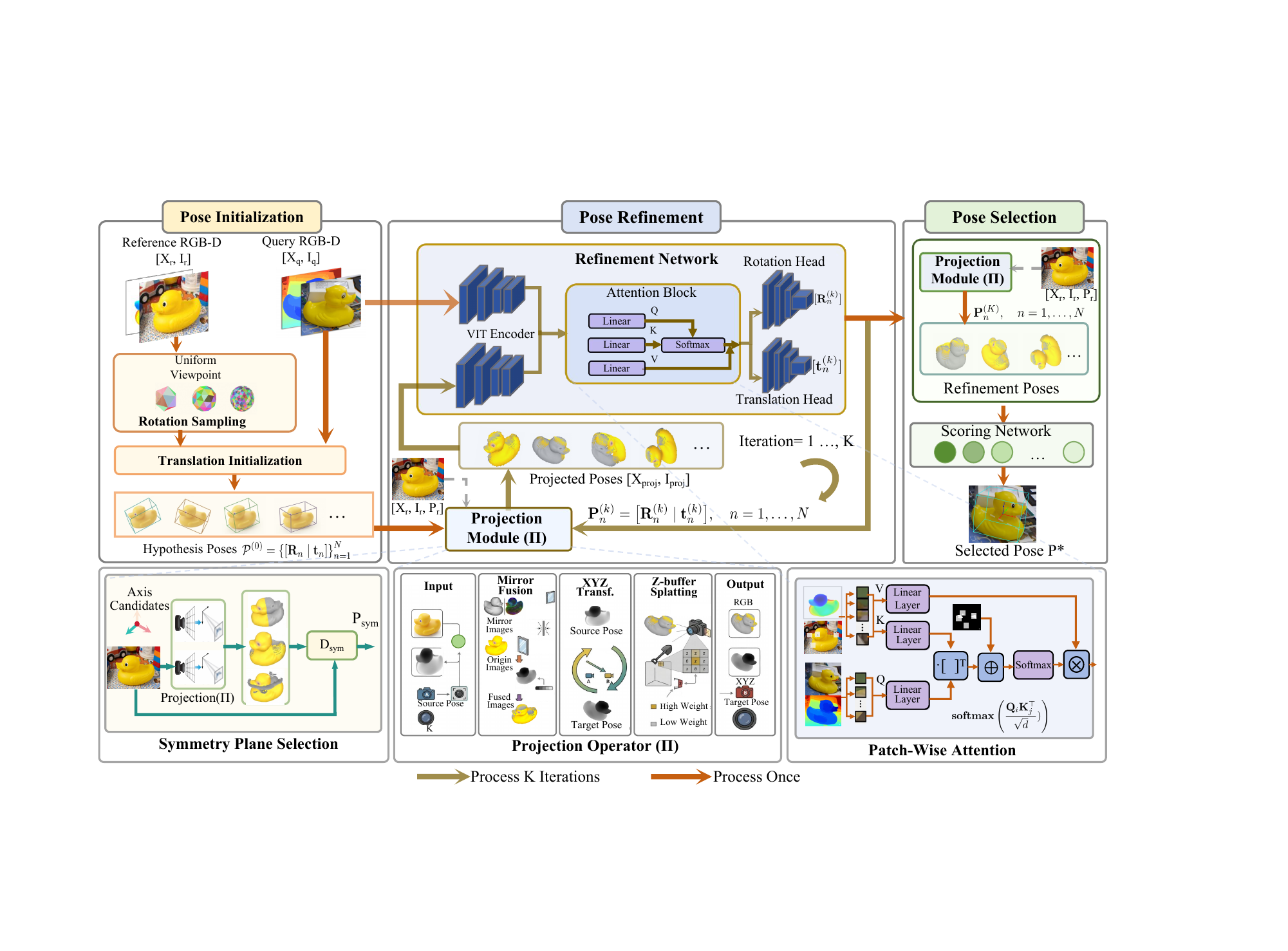}
    \captionsetup{
        font=small,
        labelfont=bf,
        justification=justified,
    }
\caption{\textbf{Overall architecture of OneViewAll.} The pipeline estimates the 6D pose through four stages:
(1) \textbf{Pose Initialization:} coarse rotation and translation hypotheses $\mathcal{P}^{(0)}$ are sampled;
(2) \textbf{Projection:} the reference RGB-D view is warped to each hypothesis via pose-conditioned projection, using symmetry-aware mirror fusion and splatting;
(3) \textbf{Pose Refinement:} each hypothesis $\mathbf{P}_n^{(k)}$ is iteratively updated over $K$ iterations by a network that aligns projected and query features via patch-wise attention;
(4) \textbf{Pose Selection:} refined hypotheses are scored, and the one with the highest score is selected as the final pose $\mathbf{P}^*$.
Bottom insets illustrate key components of the pipeline.}
\label{fig:Overview of Framework}
\end{figure*}

\subsection{Model-based 6D Pose Estimation}
The model‑based paradigm assumes that an accurate CAD model is available for each target object, providing a strong geometric prior that has long driven the state of the art. Indeed, these methods consistently occupy the top positions on benchmarks such as BOP~\cite{bop2023challenge, bop2024challenge}, and their precision sets an upper bound for the entire field. Technically, existing approaches can be broadly grouped into two lines. The first establishes 2D–3D correspondences by regressing dense coordinate maps or surface embeddings~\cite{GDR-Net_2021, SurfEmb_2022}, followed by PnP~\cite{lepetit2009epnp} and RANSAC~\cite{fischler1981ransac}. The second follows a render‑and‑compare strategy, where a CAD model is rendered under a set of pose hypotheses and the resulting templates are compared with the query image at the pixel or feature level, as exemplified by MegaPose~\cite{MegaPose} and FoundationPose~\cite{FoundationPose_2024}. Despite their outstanding accuracy, the fundamental limitation of both directions is their heavy reliance on high‑fidelity CAD models. Acquiring such models is costly and labor‑intensive, and their absence prevents the deployment in open‑world or large‑scale settings. The rendering pipeline also introduces non‑trivial computational overhead.

\subsection{Model-free 6D Pose Estimation}
To bypass the dependency on CAD models, model‑free methods capture object priors from reference images. Early approaches relied on multi‑view setups and explicit 3D reconstruction via SfM, as seen in OnePose~\cite{OnePose_2022} and OnePose++~\cite{He_2022_OnePose++}. More recent works pursue a Generation‑as‑Reconstruction strategy to approach model‑based accuracy. Any6D~\cite{Any6D_2025} employs fast reconstruction tools such as InstantMesh~\cite{xu2024instantmesh} to build meshes from sparse views, while OnePoseViaGen~\cite{geng2025oneview} reconstructs a textured mesh from a single image via a generative pipeline, using text‑guided randomization to bridge the domain gap. Although these methods achieve competitive accuracy, they rely on an intermediate 3D representation whose geometric errors (caused by single‑view ambiguity or reconstruction artifacts) remain fixed and cannot be corrected later. Some other methods that avoid explicit 3D reconstruction face different limitations. One2Any~\cite{One2Any_2025} performs single‑view feature matching, but its reliance on 2D appearance provides weak geometric constraints when the visible surface is sparse or self‑occluded. SinRef‑6D~\cite{Liu2025SinRef6D} introduces stronger geometry by aligning reference and query point clouds through SVD‑based 3D–3D matching; however, like One2Any, it operates on the partially observed geometry and lacks a mechanism to infer the invisible back side. This makes both methods susceptible to local minima and unable to resolve orientation ambiguities.

We address these issues by combining symmetry‑aware mirror fusion with direct projection‑based alignment. Mirror fusion hallucinates the unseen back‑side geometry through a self‑checked plane mechanism, providing a complete geometric context without explicit 3D reconstruction. Instead of relying on CAD models, our method directly warps the reference observation under candidate poses, avoiding the need to commit to a single intermediate 3D shape. This design keeps the iterative pipeline anchored to the original reference image.

\section{Method}
\subsection{Overview}
OneViewAll estimates the 6D pose of an object from a single reference RGB‑D view without CAD models, following a project‑and‑compare paradigm. As shown in Fig.~\ref{fig:Overview of Framework}, the pipeline consists of three stages.
Sec.~\ref{subsec:Initialization} describes how coarse pose hypotheses are generated to cover the search space.
Sec.~\ref{subsec:Projection Module} introduces the projection module that maps the reference observation to each hypothesis view, augmented by symmetry‑aware mirror fusion.
Sec.~\ref{subsec:Pose Refinement} explains the iterative refinement network that updates poses by comparing projected and query features.
Finally, Sec.~\ref{subsec:Pose Selection} presents the selection strategy that picks the best pose from the refined set.

\subsection{Formulation of Project-and-Compare:}
\label{subsec:Project-and-Compare}
The innovation of OneViewAll is the Project-and-Compare paradigm. Unlike classical Render-and-Compare, which depends on CAD models and rendering, we generalize the idea to the model-free setting. Specifically, we replace rendering with reference-conditioned projection in a projection-equivariant observation space. This creates a unified framework in which both approaches can be viewed as observation-consistency-driven pose estimation under different synthesis mechanisms.

In classical Render-and-Compare, pose estimation relies on an explicit 3D object model $\mathcal{M}$ to generate observations via rendering, which are then aligned with the query observation $\mathcal{O}_q$. In contrast, we operate directly on RGB-D observations. Each observation is represented as a geometry-aware pair $\mathcal{O}_r = [\mathbf{X}_r, \mathbf{I}_r]$, where $\mathbf{I}_r \in \mathbb{R}^{3 \times H \times W}$ is the RGB image and $\mathbf{X}_r \in \mathbb{R}^{3 \times H \times W}$ encodes 3D coordinates per-pixel obtained by depth back-projection using camera intrinsics $\mathbf{K}$. The reference pose is denoted as $\mathbf{P}_r$. Given a discrete set of pose hypotheses $\mathcal{P}^{(0)} = \{\mathbf{P}_n^{(0)}\}_{n=1}^{N}$, each hypothesis induces a geometry-aware projection of the reference observation:
\begin{equation}
[\mathbf{X}_{\text{proj}}^{(n)}, \mathbf{I}_{\text{proj}}^{(n)}] = \Pi(\mathcal{O}_r, \mathbf{P}_r, \mathbf{P}_n).
\end{equation}

We formulate refinement as predicting a relative 6D pose update that jointly corrects rotation and translation to reduce misalignment between the projected reference and the query observation. This process unifies pose estimation as iterative refinement in a projection-consistent space.

\begin{equation}
\label{eq:pose-objective}
\mathbf{P}_n^{(k+1)} = \mathcal{F}\big(\mathbf{P}_n^{(k)}, \Pi(\mathcal{O}_r,  \mathbf{P}_r, \mathbf{P}_n^{(k)}), \mathcal{O}_q\big),
\end{equation}
where $\mathcal{F}$ iteratively refines the pose based on the current projected observation and its relationship to the query observation. 

After $K$ iterations, we obtain a refined set of pose hypotheses $\{\mathbf{P}_n^{(K)}\}_{n=1}^{N}$. The final pose is selected by evaluating the consistency between the refined projections and the query observation:
\begin{equation}
\label{eq:scoring_definition}
\mathbf{P}^* = \arg\min_{n} \mathcal{C}\big(\Pi(\mathcal{O}_r, \mathbf{P}_r, \mathbf{P}_n^{(K)}), \mathcal{O}_q\big).
\end{equation}

Overall, this formulation unifies hypothesis generation, reference-conditioned projection, and iterative pose refinement into a CAD-free instantiation of the Render-and-Compare paradigm.

\subsection{Pose Initialization}
\label{subsec:Initialization}
We generate initial pose hypotheses by pairing a set of rotation candidates with a shared translation estimate. The full set of rotation hypotheses, $\mathcal{R}_{full}$, discretizes $SO(3)$ using a Fibonacci lattice for viewpoint directions $\mathcal{V}\subset S^2$ and a set of discrete in‑plane rotations $\mathcal{R}_{ip}\subset SO(2)$ about the canonical object axis $\mathbf{z}_{obj}$:
\begin{equation}
\mathcal{R}_{full} = \left\{ 
\bigl( \mathbf{R}(\mathbf{d}) \mathbf{R}_{\alpha} \bigr)^{-1} 
\;\middle|\; 
\mathbf{d} \in \mathcal{V},\; \mathbf{R}_{\alpha} \in \mathcal{R}_{ip} 
\right\}.
\end{equation}
For scenes with an upright layout (e.g., objects resting on a horizontal supporting surface), we prune $\mathcal{R}_{full}$ by thresholding the relative camera height:
\begin{equation}
\label{eq:pruned}
\mathcal{R}_{pruned} = \left\{ 
\mathbf{R} \in \mathcal{R}_{full} 
\;\middle|\; 
[\mathbf{T}_{cam}]_{z} \geq \tau
\right\},
\end{equation}
where $[\mathbf{T}_{cam}]_{z}$ denotes the $z$-translation component of the camera pose within the object coordinate system.

We establish two operating modes to balance computational efficiency and geometric coverage. The total number of raw rotation hypotheses is
\begin{equation}
N_{max} = N_{sphere} \times |\mathcal{R}_{ip}|,
\qquad 
|\mathcal{R}_{ip}| = \frac{360^\circ}{\Delta\phi},
\end{equation}
where $N_{sphere}=|\mathcal{V}|$ is the number of viewpoint directions and $\Delta\phi$ the in‑plane rotation step. Table~\ref{tab:pose_modes} summarizes the configurations. The \textit{full} mode uses the same sampling ($N_{sphere}=60$, $\Delta\phi=90^\circ$) and generates $N_{max}=240$ raw hypotheses; applying the height pruning in Eq.~\eqref{eq:pruned} yields $N=80$ (\textit{standard}), while disabling it retains all $N=240$ (\textit{full}).

\begin{table}[htbp]
\centering
\caption{Parameter settings for pose pruning ($\tau = 0.35$). The \textit{standard} mode shares the same sampling strategy and can be used with or without height pruning.}
\label{tab:pose_modes}
\setlength{\tabcolsep}{3pt}
\begin{tabular}{lccccc}
\toprule
\textbf{Mode} & $N_{sphere}$ & $\Delta\phi$ & In-plane Rots. & $N_{max}$ & Pruned $N$ \\ 
\midrule
lightweight & 20 & $180^\circ$ & 2 & 40  & 14  \\
standard & 60 & $90^\circ$  & 4 & 240 & 80 \\ 
\bottomrule
\end{tabular}
\end{table}

To accommodate different application requirements, the two modes trade off accuracy against inference speed through controlled hypothesis sparsity. The lightweight mode targets low‑latency scenarios such as real‑time tracking with small frame‑to‑frame motion; it uses $N_{sphere}=20$ and $\Delta\phi=180^\circ$, yielding $N=14$ hypotheses after height pruning. Our default standard configuration employs the standard sampling with height pruning, producing $N=80$ hypotheses and striking a balance between accuracy and efficiency. When maximum viewpoint coverage is required (e.g., unconstrained orientations with inverted or tumbling objects), the standard variant disables pruning and retains all $N=240$ hypotheses at the cost of higher inference time.

\subsection{Projection Module}
\label{subsec:Projection Module}

The reference observation consists of an RGB image $\mathbf{I}_r$ and a per‑pixel 3D point cloud $\mathbf{X}_r$ back‑projected from the reference depth using the camera intrinsics $\mathbf{K}$.
We enhance projection robustness under partial visibility by incorporating object‑level symmetry priors into the projection operator $\Pi$ shown in the bottom‑middle of Fig.~\ref{fig:Overview of Framework}.
A lightweight geometric self‑check first determines the dominant symmetry axis from the reference view; the full projection pipeline then operates on the symmetry‑augmented reference.

\textbf{1) Symmetry Plane Selection.}
When the symmetry plane $\mathcal{P}_{\mathrm{sym}}$ is unknown, we select it from a compact set of candidate planes defined in the reference object frame.
Specifically, we consider the three orthogonal coordinate planes
$\mathcal{P}_{X}(x=0)$, $\mathcal{P}_{Y}(y=0)$, and $\mathcal{P}_{Z}(z=0)$,
corresponding to reflections along the canonical axes $\{A_X,A_Y,A_Z\}$.
This design follows the annotation convention of common BOP benchmarks~\cite{bop2023challenge}, where object coordinate frames are typically aligned with dominant geometric directions. Therefore, the dominant symmetry planes of symmetric objects are well covered by this three-plane candidate set.

We also include a no-symmetry baseline. For each candidate plane, we apply the same projection pipeline used at inference, including mirror fusion, relative pose transformation, and weighted splatting, while setting the target pose to the reference pose $\mathbf{P}_r$. This yields a self-projected observation $\mathcal{D}_{\mathrm{proj}}$. A valid symmetry hypothesis should be alignment with the original reference observation $\mathcal{D}_r$ after self-projection. We measure the geometric distortion using the mean absolute error over valid reference pixels $\mathcal{V}$:
\begin{equation}
D_{\mathrm{sym}} =
\frac{1}{|\mathcal{V}|}
\sum_{p\in\mathcal{V}}
\left\|
D_{\mathrm{proj}}(p)-D_r(p)
\right\|_1 ,
\end{equation}
where $D_{\mathrm{proj}}$ and $D_r$ denote the projected and reference depths, respectively.

The plane with the smallest distortion is selected as the symmetry plane. If the minimum distortion, normalized by the object diameter, is larger than a fixed threshold $0.3$, the object is treated as asymmetric and the Mirror Fusion is disabled. While the three-plane set suffices for BOP's axis-aligned frames, unconstrained objects may benefit from predicting arbitrary symmetry planes via data-driven methods (e.g.,~\cite{huang2015single, zhou2020learning}). The predicted plane can be further verified by our self-check mechanism.

\textbf{2) Symmetry-Aware Mirror Fusion.}
\label{subsec:Symmetry-Aware Mirror Fusion}

With the symmetry axis $\mathcal{P}_{sym}$ determined in the previous step, we augment the reference observation with an hallucinated view of the back side.
The reference RGB-D observation $\mathcal{O}_r = [\mathbf{X}_r, \mathbf{I}_r]$ is first transformed into an object-centric frame through $\mathbf{P}_r$: $\mathbf{X}_{obj} = \mathbf{R}_r^\top(\mathbf{X}_r - \mathbf{t}_r)$.
Points are then reflected across $\mathcal{P}_{sym}$ to obtain a mirrored point cloud $\mathbf{X}_{mir} = \text{Reflect}(\mathbf{X}_{obj}, \mathcal{P}_{sym})$.
To provide plausible appearance for the projected geometry, we generate a mirrored texture $\mathbf{I}_{mir}$ by applying a channel-wise aggregation $\Psi$ (e.g., luminance mapping) to $\mathbf{I}_r$, which preserves the main photometric patterns while avoiding illumination inconsistencies.
Both geometry and texture are lifted back to the reference camera frame via $\mathbf{P}_r$ and concatenated with the original data along the channel dimension:
\[
\tilde{\mathbf{X}}_r = [\mathbf{X}_r, \mathbf{R}_r \mathbf{X}_{mir} + \mathbf{t}_r]_c,\quad
\tilde{\mathbf{I}}_r = [\mathbf{I}_r, \mathbf{I}_{mir}]_c.
\]
The resulting symmetry-augmented observation $\tilde{\mathcal{O}}_r = (\tilde{\mathbf{X}}_r, \tilde{\mathbf{I}}_r)$ thus encodes both the visible and the mirrored back-side geometry in a unified representation, which is subsequently used by the projection operator $\Pi$.

\textbf{3) Relative Pose Transformation.}
For each pose hypothesis $\mathbf{P}_n^{(k)} \in \mathcal{P}^{(k)}$ at iteration $k$, we transform the mirror-fused observation $\tilde{\mathcal{O}}_r = (\tilde{\mathbf{X}}_r, \tilde{\mathbf{I}}_r)$ into the corresponding pose-aligned coordinate frame. We treat $\tilde{\mathbf{X}}_r$ as a dense pixel-aligned 3D field, where each sample is denoted as a point $\mathbf{p}_i \equiv \tilde{\mathbf{X}}_r(u_i, v_i)$ with associated appearance $\mathbf{c}_i \equiv \tilde{\mathbf{I}}_r(u_i, v_i)$. Here, $i = 1, \dots, |\tilde{\mathbf{X}}_r|$ indexes all valid pixels in the symmetry-augmented reference observation, covering both the original visible surface and the mirrored back-side points. Each point is first expressed in the reference camera frame and then re-mapped under the hypothesis pose $\mathbf{P}_n^{(k)} = [\mathbf{R}_n^{(k)} \mid \mathbf{t}_n^{(k)}]$, where $\mathbf{R}_n^{(k)} \in SO(3)$ and $\mathbf{t}_n^{(k)} \in \mathbb{R}^3$ denote the rotation and translation of the $n$-th pose hypothesis at iteration $k$:
\begin{equation}
\mathbf{p}_i^{(n,k)} = \mathbf{R}_n^{(k)} \mathbf{R}_r^\top (\mathbf{p}_i - \mathbf{t}_r) + \mathbf{t}_n^{(k)}.
\end{equation}
This induces a pose-conditioned transformed representation:
\begin{equation}
\tilde{\mathcal{O}}_{trans}^{(n,k)} = \{ (\mathbf{p}_i^{(n,k)}, \mathbf{c}_i) \}_{i=1}^{|\tilde{\mathbf{X}}_r|},
\end{equation}
where $\mathbf{c}_i$ denotes appearance attributes (RGB or grayscale), which remain invariant under rigid SO(3) transformations.
\textbf{4) Weighted Z-buffer Splatting:}

After pose transformation, each point $\mathbf{p}_i^{(n,k)}=(x_i,y_i,z_i)^\top$ projects to continuous image coordinates $(x_i,y_i)$. To resolve visibility when multiple points (including symmetry-mirrored ones) fall on the same pixel, we adopt Z-buffer point splatting~\cite{Zwicker2001, Kerbl2023}. Every point contributes to a $3\times3$ neighborhood. The visibility energy for pixel $(u,v)$ is
\begin{equation}
E_i^{(n,k)}(u,v) = z_i + \lambda\,\mathcal{K}(u-x_i,\;v-y_i),
\end{equation}
with a fixed kernel
\begin{equation}
\mathcal{K} = \begin{bmatrix}
0.002 & 0.001 & 0.002\\
0.001 & 0     & 0.001\\
0.002 & 0.001 & 0.002
\end{bmatrix},
\end{equation}
and $\lambda=1$. The kernel acts as a local regularizer; its maximum perturbation ($2\times10^{-3}\,\mathrm{m}$) is negligible compared to typical object depths ($0.3$--$1.0\,\mathrm{m}$), thus preserving depth ordering while reducing discretization holes. Visibility is decided by the standard Z-buffer:
\begin{equation}
i^*(u,v) = \arg\min_i E_i^{(n,k)}(u,v),
\end{equation}
selecting the nearest surface along each ray. Mirror fusion adds surface hypotheses without overriding originals; all points compete equally. The index map then retrieves geometry and appearance:
\begin{equation}
\left[ \mathbf{X}_{\mathrm{proj}}^{(n,k)}(u,v), \mathbf{I}_{\mathrm{proj}}^{(n,k)}(u,v) \right] = \tilde{\mathcal{O}}_{\mathrm{trans}}^{(n,k)}\!\left[i^*(u,v)\right].
\end{equation}
This weighted Z-buffer competition and index-map retrieval form a regularized splatting operator that yields dense, visibility-consistent projections with suppressed discretization artifacts.

\subsection{Pose Refinement}
\label{subsec:Pose Refinement}

Given a initial or tracked pose hypothesis $\mathbf{P}_n^{(k)}$ at iteration $k$, we utilize a projection module $\Pi$ to render a pose-conditional reference view $[\mathbf{X}_{\text{proj}}^{(n,k)}, \mathbf{I}_{\text{proj}}^{(n,k)}] = \Pi(\tilde{\mathcal{O}}_r, \mathbf{P}_r, \mathbf{P}_n^{(k)})$. Both the query observation and this projected reference view are encoded via a shared CNN-ViT backbone~\cite{CvT_2021, mehta2022mobilevit}, chosen for its favorable balance between accuracy and efficiency, and output patch-level geometric-appearance tokens $\mathbf{f}_q^\prime$ and $\mathbf{f}_r^{\prime(n,k)}$.

A cross-attention mechanism then aggregates these features by setting the queries $Q = \mathbf{f}_q^\prime$ and keys/values $K,V = \mathbf{f}_r^{\prime(n,k)}$:
\begin{equation}
\mathcal{A}^{(n,k)} = \mathrm{softmax}\left( \frac{\mathbf{f}_q^\prime (\mathbf{f}_r^{\prime(n,k)})^\top}{\sqrt{d}} \right) \mathbf{f}_r^{\prime(n,k)}.
\end{equation}
The attended feature tokens are fed into a linear regression head to predict the relative pose updates $(\Delta \mathbf{R}_n^{(k)}, \Delta \mathbf{t}_n^{(k)})$.

\textbf{Training and Loss.} The network is optimized using the Adam optimizer ($\text{lr}=10^{-4}$) on the RGB-D data from GSO~\cite{downs2022google} and ModelNet~\cite{wu20153d} to ensure category-agnostic generalization. For a ground-truth pose $\mathbf{P}^\star = [\mathbf{R}^\star \mid \mathbf{t}^\star]$, the target relative transformations are formulated as $\Delta \mathbf{R}_n^\star = \mathbf{R}^\star (\mathbf{R}_n^{(k)})^\top$ and $\Delta \mathbf{t}_n^\star = \mathbf{t}^\star - \mathbf{t}_n^{(k)}$. We minimize the joint regression loss:
\begin{equation}
\begin{aligned}
\mathcal{L}_{\text{refine}} = \mathbb{E}_{n,k} \Big[ & \bigl\| \Delta \mathbf{t}_n^{(k)} - \Delta \mathbf{t}_n^\star \bigr\|_2 \\
& + \bigl\| \Delta \mathbf{R}_n^{(k)} - \Delta \mathbf{R}_n^\star \bigr\|_F \Big].
\end{aligned}
\end{equation}

\subsection{Pose Selection}
\label{subsec:Pose Selection}

After $K$ refinement iterations, we obtain the final candidate set $\{\mathbf{P}_n^{(K)}\}_{n=1}^{N}$. We reuse the CNN-ViT encoder and the attention block to extract global features from both the query and the hypothesis-conditioned projections $[\mathbf{X}_{\text{proj}}^{(n,K)}, \mathbf{I}_{\text{proj}}^{(n,K)}]$. The concatenated embedding of each pair is processed by a MLP to output a scalar confidence logit $s_n$. The final prediction is determined via $\hat{\mathbf{P}} = \mathbf{P}_{n^*}^{(K)}$, where $n^* = \arg\max(s_n)$, balancing rich global context with computational efficiency.

Following multi-hypothesis distribution matching practices~\cite{FoundationPose_2024}, we optimize the scoring MLP over the entire hypothesis space using a Softmax Cross-Entropy loss. We first define the geometric residual error $\mathcal{E}_n$ for each candidate:
\begin{equation}
\mathcal{E}_n = \|\log(\mathbf{R}^\star (\mathbf{R}_n^{(K)})^\top)\|_F + \|\mathbf{t}_n^{(K)} - \mathbf{t}^\star\|_2.
\end{equation}
The target probability $p_n^\star$ and predicted probability $p_n$ for the $n$-th hypothesis are implicitly coupled and normalized via:
\begin{equation}
p_n^\star = \frac{\exp(-\mathcal{E}_n / \sigma)}{\sum_{j=1}^N \exp(-\mathcal{E}_j / \sigma)}, \qquad p_n = \frac{\exp(s_n)}{\sum_{j=1}^N \exp(s_j)},
\end{equation}
where $\sigma$ is a temperature hyperparameter. The confidence verification head is supervised by minimizing the cross-entropy objective:
\begin{equation}
\mathcal{L}_{\text{conf}} = -\sum_{n=1}^{N} p_n^\star \ln p_n.
\end{equation}
This global alignment formulation forces the scoring network to implicitly learn mathematical quality rankings while suppressing far-outlier negative proposals.

\definecolor{highlightblue}{RGB}{235, 235, 255}

\begin{table*}[t]
\centering
\small
\captionsetup{
    font=small,
    labelfont=bf,
    justification=justified,
}
\caption{Comparison of test results on the LINEMOD dataset (ADD-0.1d \%). Our method is evaluated with both rendered reference images ($^{\dagger}$) and real-world reference views ($^{*}$). Labels such as $N$ indicate the number of hypotheses.}
\label{tab:linmod}
\setlength{\tabcolsep}{1.8pt} 
\renewcommand{\arraystretch}{1.3} 
\begin{tabular}{lcccccccccccccccccc}
\toprule
\multirow{2}{*}{Method} & \multirow{2}{*}{Year} & \multirow{2}{*}{Mod.} & \multirow{2}{*}{\begin{tabular}[c]{@{}c@{}}Ref.\\ Num.\end{tabular}} & \multicolumn{13}{c}{Object ID} & \multirow{2}{*}{\textbf{Mean}} & \multirow{2}{*}{\begin{tabular}[c]{@{}c@{}}Time\\ (ms)\end{tabular}} \\ 
\cmidrule(lr){5-17}
 & & & & ape & bench & cam & can & cat & driller & duck & eggbox & glue & holep. & iron & lamp & phone & & \\ 
\midrule
OnePose*~\cite{OnePose_2022} & 2022 & RGB & 200 & 11.8 & 92.6 & 88.1 & 77.2 & 47.9 & 74.5 & 34.2 & 71.3 & 37.5 & 54.9 & 89.2 & 87.6 & 60.6 & 63.6 & 66 \\
OnePose++*\cite{He_2022_OnePose++} & 2023 & RGB & 200 & 31.2 & 97.3 & 88.0 & 89.8 & 70.4 & 92.5 & 42.3 & 99.7 & 48.0 & 69.7 & 97.4 & 97.8 & 76.0 & 76.9 & 88 \\
LatentFusion*~\cite{LatentFusion_2020} & 2020 & RGB-D & 16 & 88.0 & 92.4 & 74.4 & 88.9 & 94.5 & 91.7 & 68.1 & 96.3 & 49.4 & 82.1 & 74.6 & 94.7 & 91.5 & 83.6 & -- \\
FS6D + ICP*~\cite{FS6D_2022} & 2022 & RGB-D & 16 & 78.0 & 88.5 & 91.0 & 89.5 & 97.5 & 92.0 & 75.5 & 99.5 & 99.5 & 96.0 & 87.5 & 97.0 & 97.5 & 91.5 & 185 \\
FS6D*~\cite{FS6D_2022} & 2022 & RGB-D & 16 & 74.0 & 86.0 & 88.5 & 86.0 & 98.5 & 81.0 & 68.5 & 100.0 & 99.5 & 97.0 & 92.5 & 85.0 & 99.0 & 88.9 & 72 \\
iG-6DoF*~\cite{iG-6DoF_2025} & 2025 & RGB & 16
 & 64.3 & 96.3 & 88.6 & 92.1 & 83.2 & 88.6 & 73.3 & 99.6 & 81.3 & 94.3 & 81.3 & 88.6 & 73.1 & 85.1 & 500 \\
NOPE*~\cite{NOPE_2024} & 2024 & RGB & 1 + GT & 2.0 & 4.5 & 2.5 & 2.2 & 0.7 & 4.7 & 0.5 & 100.0 & 79.4 & 2.9 & 4.5 & 4.2 & 3.9 & 16.3 & 1190 \\
Oryon*~\cite{Oryon_2024} & 2024 & RGB-D & 1 & 1.2 & 1.3 & 3.9 & 0.8 & 12.7 & 8.5 & 0.8 & 63.2 & 18.4 & 1.6 & 0.6 & 2.9 & 11.7 & 9.8 & 900 \\
One2Any*~\cite{One2Any_2025} & 2025 & RGB-D & 1 & 33.1 & 15.7 & 72.7 & 37.0 & 66.2 & 68.2 & 35.8 & 100.0 & 99.9 & 42.0 & 28.2 & 31.9 & 53.2 & 52.6 & 90 \\
SinRef-6D$^{\dagger}$~\cite{Liu2025SinRef6D} & 2025 & RGB-D & 1 & 85.7 & 99.3 & 73.2 & 98.3 & 93.0 & 98.7 & 66.6 & 98.5 & 99.1 & 74.6 & 90.9 & 97.6 & 97.4 & 90.8 & -- \\
\midrule
\rowcolor{highlightblue}
\textbf{Ours* ($N=14$)} & \textbf{2026} & \textbf{RGB-D} & \textbf{1} & \textbf{61.6} & \textbf{98.9} & \textbf{84.0} & \textbf{94.3} & \textbf{95.9} & \textbf{92.9} & \textbf{96.4} & \textbf{85.1} & \textbf{94.5} & \textbf{95.9} & \textbf{98.5} & \textbf{93.3} & \textbf{76.5} & \textbf{89.8} & \textbf{80} \\
\rowcolor{highlightblue}
\textbf{Ours* ($N = 80$)} & \textbf{2026} & \textbf{RGB-D} & \textbf{1} & \textbf{62.1} & \textbf{98.9} & \textbf{84.1} & \textbf{93.7} & \textbf{95.3} & \textbf{94.8} & \textbf{96.3} & \textbf{99.92} & \textbf{98.8} & \textbf{96.3} & \textbf{98.3} & \textbf{93.4} & \textbf{81.9} & \textbf{91.8} & \textbf{375} \\
\rowcolor{highlightblue}
\textbf{Ours$^{\dagger}$ ($N=14$)} & \textbf{2026} & \textbf{RGB-D} & \textbf{1} & \textbf{88.3} & \textbf{96.9} & \textbf{89.8} & \textbf{94.2} & \textbf{96.3} & \textbf{94.3} & \textbf{95.0} & \textbf{79.7} & \textbf{95.1} & \textbf{92.2} & \textbf{95.3} & \textbf{89.1} & \textbf{80.8} & \textbf{91.2} & \textbf{80} \\
\rowcolor{highlightblue}
\textbf{Ours$^{\dagger}$ ($N = 80$)} & \textbf{2026} & \textbf{RGB-D} & \textbf{1} & \textbf{93.2} & \textbf{99.8} & \textbf{91.1} & \textbf{99.7} & \textbf{99.9} & \textbf{100.0} & \textbf{99.4} & \textbf{100.0} & \textbf{100.0} & \textbf{98.1} & \textbf{100.0} & \textbf{100.0} & \textbf{94.5} & \textbf{98.8} & \textbf{375} \\
\bottomrule   
\end{tabular}
\end{table*}

\section{EXPERIMENTS}
\subsection{Datasets and Evaluation Metrics}

We evaluate OneViewAll under the strict single-reference RGB-D setting on multiple challenging benchmarks. We report results using both rendered references and real-world references where applicable.
LINEMOD \& LM-O~\cite{hinterstoisser2012linemod, bop2024challenge}: We use the 15 texture-less objects from LINEMOD for base evaluation and its occluded subset LM-O to test robustness against heavy occlusion. Both rendered and real-world references are used. We report ADD-0.1d accuracy for LINEMOD and BOP Average Recall (AR) for LM-O.

Real275~\cite{real275_2019} and Toyota-Light~\cite{hodan2018bop}: These datasets focus on generalization to unseen object instances and challenging illumination. We use real-world references and report BOP AR.

The metrics are defined as follows:
\begin{enumerate}
    \item \textbf{ADD-0.1d}: A pose is considered correct if the average vertex distance $e_{\text{ADD}}$ is less than 10\% of the object diameter $d$:
    \begin{equation}
        e_{\text{ADD}} = \frac{1}{|\mathcal{M}|} \sum_{x \in \mathcal{M}} \|(Rx + T) - (R^*x + T^*)\|_2
    \end{equation}
    The \textbf{ADD-0.1} accuracy represents the percentage of test samples where $e_{\text{ADD}} < 0.1d$. 
    \item \textbf{ADD(-S) AUC}: Calculated as the area under the accuracy-threshold curve (from 0 to 10cm). For symmetric objects, $e_{\text{ADD-S}}$ uses the distance to the closest vertex to account for rotational ambiguity.
    \item \textbf{BOP AR}: The arithmetic mean of three symmetry-aware scores: Visible Surface Discrepancy (VSD), Maximum Surface Distance (MSSD), and Maximum Projection Distance (MSPD):
    \begin{equation}
        \text{AR} = \frac{1}{3} (\text{AR}_{\text{VSD}} + \text{AR}_{\text{MSSD}} + \text{AR}_{\text{MSPD}})
    \end{equation}
\end{enumerate}

\subsection{Implementation Details}

OneViewAll is implemented in PyTorch and executed on a single NVIDIA RTX 4090 GPU. 

\textbf{1) Evaluation Protocol.}
This work follows a single-reference setting with no CAD model access during inference. 
The primary results use one real-world RGB-D reference per object. 
We additionally report results using rendered references for controlled comparisons with prior methods that adopt this practice. 
Rendered references serve two purposes.
They enable a fair comparison with methods like SinRef-6D. 
They also provide an ideal noise-free setting for analyzing our projection pipeline.
We do not use CAD models, explicit 3D reconstruction, or multi-view data during training or inference.

\textbf{2) Reference Data Acquisition.}
A key challenge in single-reference model-free 6D pose estimation is selecting an informative reference viewpoint. In our setting, we use exactly one RGB-D image per object, which can be either a real-world image or a rendered image. For viewpoint selection, we choose the reference viewpoint that aligns with the typical pose distribution in real-world scenarios, deliberately avoiding non-informative perspectives (e.g., the bottom of the object) to capture the most discriminative visible features.

\textbf{3) Viewpoint Initialization and Pruning.}
Given the single reference image, we generate an initial set of pose hypotheses by sampling over the view sphere. On the Linemod dataset, we apply the viewpoint pruning strategy introduced in Section~\ref{subsec:Initialization} and evaluate both the \textit{lightweight} and \textit{standard} configurations, whose parameter settings for upright scenes are provided in Table~\ref{tab:pose_modes}. For all other datasets, unless stated otherwise, we adopt the \textit{standard} mode without viewpoint pruning, using the raw set of initial views.

\begin{figure}[t!]
    \centering
    \includegraphics[trim={3cm 0cm 4cm 0cm}, clip, width=1\columnwidth]{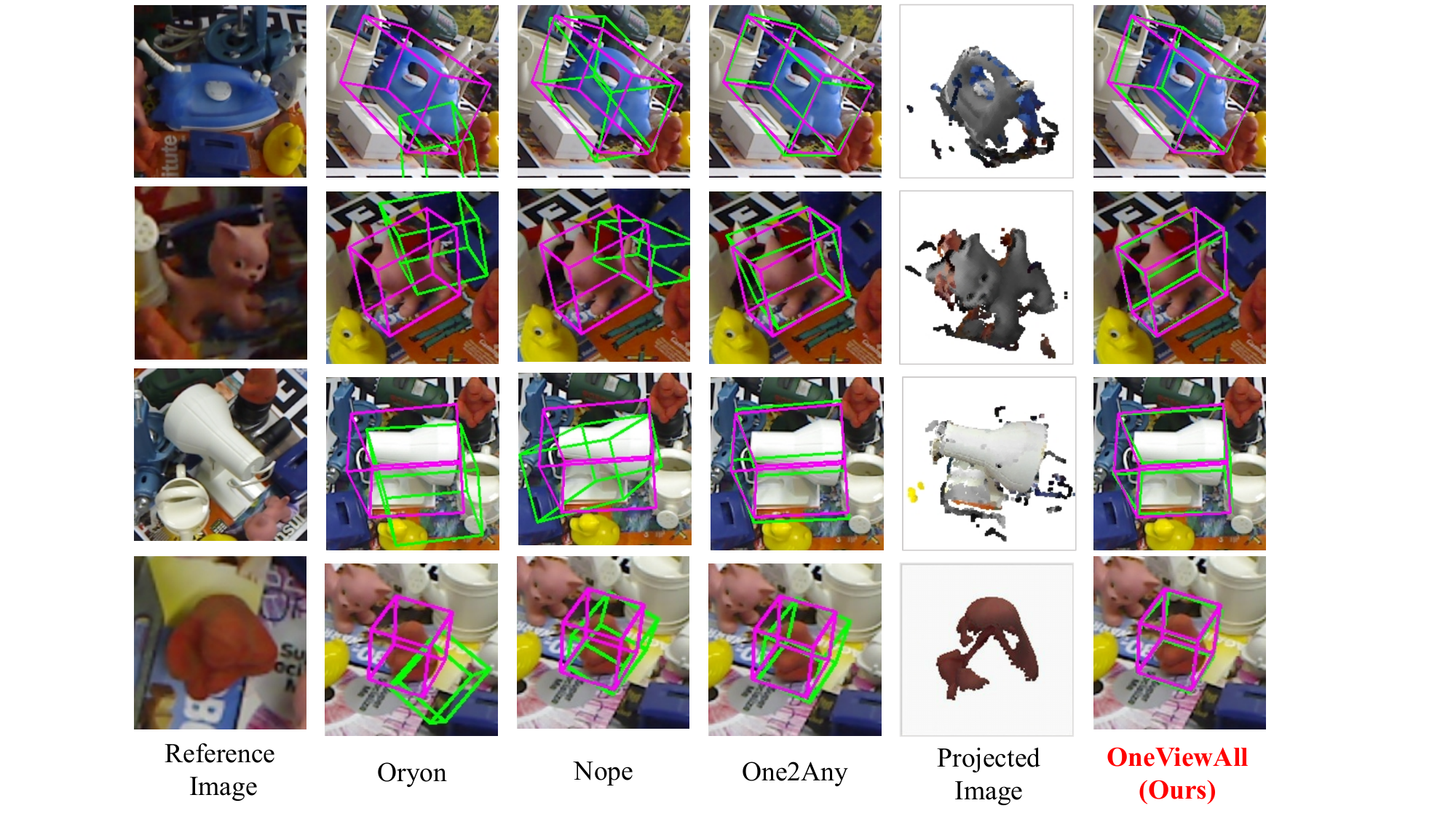}
    \captionsetup{
        font=small,
        labelfont=bf,
        justification=justified,
    }
    \caption{Qualitative results on LINEMOD using real reference images. Red and green boxes denote ground-truth and predicted poses. Compared with Oryon~\cite{Oryon_2024}, NOPE~\cite{NOPE_2024}, and One2Any~\cite{One2Any_2025}, our method achieves more accurate alignments. The projected image shows results with symmetry-aware mirror fusion.}
    \label{fig:linemod}
\end{figure}

\subsection*{C. Quantitative Comparisons}

\textbf{LINEMOD.} Table~\ref{tab:linmod} compares OneViewAll with prior model‑free methods~\cite{OnePose_2022, He_2022_OnePose++, LatentFusion_2020, FS6D_2022, iG-6DoF_2025, NOPE_2024, Oryon_2024, One2Any_2025, Liu2025SinRef6D}. Note that SinRef‑6D~\cite{Liu2025SinRef6D} uses rendered reference images ($^{\dagger}$) rather than real‑world ones, and such rendered images are generated from CAD models; we therefore evaluate OneViewAll under both modalities to isolate the effect of reference type. Our method with real‑world references (Ours*, $N=80$) already reaches 91.8\% ADD‑0.1d, surpassing SinRef‑6D's 90.8\% without relying on any CAD‑derived information. When evaluated with rendered references for a identical‑modality comparison, OneViewAll$^{\dagger}$ ($N=80$) achieves 98.8\%, substantially higher than SinRef‑6D. The rendered‑reference results are, as expected, better than the real‑world ones because rendered images are noise‑free and benefit from perfect CAD geometry, yet the fact that Ours* already exceeds a method that implicitly exploits CAD models demonstrates the robustness of our projection‑based framework and symmetry‑aware mirror fusion under realistic sensing conditions. As $N$ increases from 14 to 80, accuracy improves from 89.8\% to 91.8\% while inference time rises from 80\,ms to 375\,ms, offering a favorable accuracy–efficiency trade‑off.

\definecolor{highlightblue}{RGB}{235, 235, 255}

\begin{table}[t]
    \centering
    \small
    \renewcommand{\arraystretch}{1.3} 
      \captionsetup{
        font=small,
        labelfont=bf,
        justification=justified,
    }
   \caption{Model-free pose estimation results (AUC of AR, MSPD, MSSD, VSD) on the LM-O dataset using a single real-world reference image. All methods utilize CNOS~\cite{CNOS_2023} for initial object segmentation.}
   \label{tab:lmo_results}
    \setlength{\tabcolsep}{1pt} 
    \begin{tabular}{l cccccc}
        \toprule
        \textbf{Method} & \textbf{Year} & \textbf{Image-to-3D} & \textbf{AR (\%)} & \textbf{MSPD} & \textbf{MSSD} & \textbf{VSD} \\ 
        \midrule
        GigaPose & 2024 & Wonder3D~\cite{Wonder3D_2024} & 17.5 & 35.8 & 9.0 & 7.6 \\
        Any-6D & 2025 & Wonder3D & 28.6 & 36.1 & 32.0 & 17.6 \\
        Any-6D & 2025 & InstantMesh~\cite{xu2024instantmesh} & 25.2 & 29.5 & 27.4 & 18.7 \\
        \rowcolor{highlightblue}
        \textbf{Ours (N = 240)} & 2026 & N/A & \textbf{40.7} & \textbf{47.3} & \textbf{43.4} & \textbf{31.6} \\
        \bottomrule
    \end{tabular}
\end{table}

\definecolor{highlightblue}{RGB}{235, 235, 255}

\begin{table}[htbp]
    \centering
    \small
    \captionsetup{
        font=small,
        labelfont=bf,
        justification=justified,
    }
   \caption{Comparison of pose estimation results on Real275 and Toyota-Light datasets using single real-world reference images. AR denotes the average recall.}
    \label{tab:real_toyota_results}
    \setlength{\tabcolsep}{4pt} 
    \renewcommand{\arraystretch}{1.3}
    \begin{tabular}{l l c c c}
        \toprule
        \textbf{Dataset} & \textbf{Method} & \textbf{Year} & \textbf{Modality} & \textbf{AR (\%)} \\
        \midrule
        \multirow{7}{*}{Real275} 
        & PoseDiffusion~\cite{PoseDiffusion_2024} & 2024 & RGB & 9.2 \\
        & RelPose++~\cite{RelPose++_2024}     & 2023 & RGB & 22.8 \\
        & ObjectMatch~\cite{ObjectMatch_2023}   & 2023 & RGBD & 26.0 \\
        & Oryon~\cite{Oryon_2024}         & 2024 & RGBD & 46.5 \\
        & One2Any~\cite{One2Any_2025}       & 2025 & RGBD & 54.9 \\
        & Any6D~\cite{Any6D_2025}        & 2025 & RGBD & 51.0 \\
        \rowcolor{highlightblue}
        & \textbf{Ours (N = 240)} & 2026 & \textbf{RGBD} & \textbf{60.1} \\
        \midrule
        --- & --- & --- & --- & --- \\
        \midrule
        \multirow{8}{*}{Toyota-Light} 
        & PoseDiffusion~\cite{PoseDiffusion_2024} & 2024 & RGB & 7.8 \\
        & RelPose++~\cite{RelPose++_2024}     & 2023 & RGB & 30.9 \\
        & ObjectMatch~\cite{ObjectMatch_2023}   & 2023 & RGBD & 9.8 \\
        & Oryon~\cite{Oryon_2024}            & 2024 & RGBD & 34.1 \\
        & One2Any~\cite{One2Any_2025}      & 2025 & RGBD & 42.0 \\
        & Any6D~\cite{Any6D_2025}          & 2025 & RGBD & 43.3 \\
        & OnePoseViaGen\cite{geng2025oneview} & 2025 & RGBD & 35.1 \\
        \rowcolor{highlightblue}
        & \textbf{Ours (N = 240)} & 2026 & \textbf{RGBD} & \textbf{50.4} \\
        \bottomrule
    \end{tabular}
\end{table}

\begin{table*}[t]
\centering
\caption{\textbf{Ablation Study of Mirror Fusion on LINEMOD.} We report the ADD-0.1d accuracy (\%) for each category. $\mathcal{P}_{sym}$ denotes the symmetry plane used in Mirror Fusion.}
\label{tab:symmetry_ablation_final}
\small
\setlength{\tabcolsep}{0.0pt}
\captionsetup{
    font=small,
    labelfont=bf,
    justification=justified,
}
\begin{tabularx}{\linewidth}{@{}l|*{13}{>{\centering\arraybackslash}X}|c@{}}
\toprule
\textbf{ID} & 1 & 2 & 4 & 5 & 6 & 8 & 9 & 10 & 11 & 12 & 13 & 14 & 15 & \multirow{2}{*}{\textbf{Mean}} \\
\textbf{Obj.} & ape & bnch. & cam & can & cat & drl. & duck & egg. & glue & hlp. & iron & lamp & phn. & \\
\midrule
\textbf{Sym.} & Asy. & Sym. & Sym. & Sym. & Sym. & Sym. & Sym. & Sym. & Sym. & Sym. & Sym. & Sym. & Sym. & - \\
\textbf{$\mathcal{P}_{sym}$} & / & $Y$ & / & $X$ & $X$ & $Y$ & $Y$ & $X$ & $X$ & $Y$ & $Y$ & $Y$ & $/$ & - \\
\midrule
\textbf{w/o} & 93.2 & 96.9 & 91.1 & 96.0 & 82.1 & 91.0 & 87.0 & 96.6 & 73.9 & 90.6 & 92.4 & 94.0 & 94.5 & 91.4 \\
\textbf{Full} & 93.2 & 99.8 & 91.1 & 99.7 & 99.9 & 100.0 & 99.4 & 100.0 & 100.0 & 98.1 & 100.0 & 100.0 & 94.5 & 98.8 \\
\bottomrule
\end{tabularx}
\end{table*}

\textbf{LM-O.} Table~\ref{tab:lmo_results} summarizes the results on the LINEMOD Occlusion (LM-O) dataset under different detection settings. With CNOS segmentation, OneViewAll achieves a mean AR of 40.7\%, which surpasses existing model-free baselines like GigaPose (17.5\%) and Any-6D (28.6\%).

\textbf{Real275 \& Toyota-Light.} In Table~\ref{tab:real_toyota_results}, we further demonstrate the generalization capability of OneViewAll on the Real275 and Toyota-Light datasets. On Real275, our method achieves 60.1\% AR, outperforming the previous best model-free method One2Any (54.9\%). On the more challenging Toyota-Light dataset, OneViewAll reaches 50.4\% AR, providing a 16\% improvement over Any-6D (43.3\%) and 43\% over the performance of OnePoseViaGen (35.1\%).

\begin{table}[ht]
\centering
\captionsetup{
    font=small,
    labelfont=bf,
    justification=justified,
}
\caption{Refinement accuracy (ADD-0.1d \%) across successive iterations. The best performance of OneViewAll is achieved at Iter. 3.}
\renewcommand{\arraystretch}{1.3} 
\label{tab:ablation_iteration}
\setlength{\tabcolsep}{5.0pt}
\begin{tabular}{lccccc}
\toprule
OneViewAll Variant & Iter. 1 & Iter. 2 & Iter. 3 & Iter. 4 & Iter. 5 \\
\midrule
Refinement with CNN  & 89.2 & 93.8 & 95.3 & 95.2 & 95.2 \\
Refinement with CNN-ViT & 92.3 & 96.7 & 98.8 & 98.7 & 98.7 \\
\bottomrule
\end{tabular}
\end{table}

\subsection*{E. Ablation Studies}

To validate the contribution of each module in OneViewAll, we conduct ablation studies on the LINEMOD dataset using rendered reference images under the \textit{single-reference RGB-D + no CAD model} setting. All variants share the same backbone and number of hypotheses as the standard model unless otherwise specified. We report mean ADD-0.1d accuracy (\%) and average inference time (ms) on a single RTX 4090 GPU.

We analyze the three components of our framework: Pose Pruning, Mirror Fusion, and CNN-ViT based Refinement.

This enables a practical balance between speed and precision.

\textbf{Mirror Fusion.}
We evaluate the symmetry-aware mirror fusion mechanism by removing the symmetry hypothesis and the mirrored back-side geometry from the projection module. As reported in Table~\ref{tab:symmetry_ablation_final}, disabling Mirror Fusion reduces mean ADD-0.1d accuracy from 98.8\% to 91.4\%, an absolute degradation of 7.4\%. The drop is especially severe on objects with approximate or full symmetry (e.g., \textit{eggbox}: 96.6\% \(\rightarrow\) 100.0\%; \textit{glue}: 73.9\% \(\rightarrow\) 100.0\%; \textit{cat}: 82.1\% \(\rightarrow\) 99.9\%). These results demonstrate that the hallucinated back-side geometry provides a complete geometric context, effectively resolving orientation ambiguity under large viewpoint changes and occlusions without additional computation.

\textbf{CNN-ViT based Refinement.}
To assess the refinement module, we conduct two controlled comparisons. First, we remove the patch-wise cross-attention and regress pose updates from pooled CNN-ViT features. This variant drops ADD-0.1d accuracy by 3.5\%, confirming that the attention mechanism helps resolve local ambiguities by focusing on discriminative geometric patches.
Second, we replace the CNN-ViT backbone with a standard CNN network while retaining the cross-attention structure. This pure CNN variant underperforms our hybrid backbone by 3.5\%, demonstrating that the ViT component based refinemet is more superior. As summarized in Table~\ref{tab:ablation_iteration}, all variants benefit from iterative refinement, but our full model remains more stable beyond three iterations, while the ablated versions tend to plateau or slightly degrade. These results validate both the patch-wise attention design and the choice of a CNN-ViT backbone for accurate and efficient pose refinement.

\section{Conclusion}
We present OneViewAll, a projection-based framework that estimates 6D pose from a single RGB‑D reference view without CAD models or explicit 3D reconstruction. It directly warps the reference observation to candidate viewpoints and aligns them with the query via iterative refinement.
A self-checked symmetry prior hallucinates invisible back-side geometry through mirror fusion, improving robustness under large viewpoint changes and occlusions.
An optional pruning step removes implausible hypotheses at initialization, further improving efficiency. Our experiments show that OneViewAll achieves 91.8\% ADD‑0.1 accuracy on LINEMOD using only one real reference view, also with consistent improvements on LM-O, Real275, and Toyota‑Light.
In tracking scenarios where inter‑frame motion is small, the refined pose from the previous frame can serve as the sole hypothesis, incurring minimal latency. This makes OneViewAll attractive for time‑sensitive applications such as robotic grasping and augmented reality. Code is available at: \url{https://github.com/tilaba/OneViewAll.git}.
\bibliographystyle{IEEEtran}
\bibliography{ref}

\end{document}